\documentclass{article}
\usepackage{amsfonts}
\usepackage{spconf,amsmath,graphicx}

\usepackage{enumitem}
\setlist{nosep, leftmargin=14pt}

\usepackage{mwe} 

\title{Unbalanced Optimal Transport for Robust Longitudinal Lesion Evolution with Registration-Aware and Appearance-Guided Priors}
\name{Melika Qahqaie\textsuperscript{1,2}, Dominik Neumann\textsuperscript{2}, Tobias Heimann\textsuperscript{2}, Andreas Maier\textsuperscript{1}, Veronika A. Zimmer\textsuperscript{2}}
\address{\textsuperscript{1} FAU Erlangen-N\"{u}rnberg  \textsuperscript{2} Siemens Healthineers, Digital Technology and Innovation, Erlangen, Germany}

\begin{document}
\maketitle


\begin{abstract}
Evaluating lesion evolution in longitudinal CT scans of cancer patients is essential for assessing treatment response, yet establishing reliable lesion correspondence across time remains challenging. Standard bipartite matchers, which rely on geometric proximity, struggle when lesions appear, disappear, merge, or split. 
We propose a registration-aware matcher based on unbalanced optimal transport (UOT) that accommodates unequal lesion mass and adapts priors to patient-level tumor-load changes. Our transport cost blends (i) size-normalized geometry, (ii) local registration trust from the deformation-field Jacobian, and (iii) optional patch-level appearance consistency. The resulting transport plan is sparsified by relative pruning, yielding one-to-one matches as well as new, disappearing, merging, and splitting lesions without retraining or heuristic rules. 
On longitudinal CT data, our approach achieves consistently higher edge-detection precision and recall, improved lesion-state recall, and superior lesion-graph component F1 scores versus distance-only baselines.

\end{abstract}


\section{Introduction}
Longitudinal assessment of tumor burden is an essential component of treatment response evaluation in cancer patients undergoing Computed Tomography (CT). Tumor burden is typically quantified by the number, size, and evolution of individual lesions within organs of interest, which requires establishing reliable correspondences between lesions across time. Identifying these longitudinal lesion matches is intrinsically difficult when lesions \emph{appear, disappear, merge, or split}, and even small registration inaccuracies or anatomical changes can disrupt geometric proximity cues used by standard matchers. Accurate detection of merging and splitting lesions is also crucial for reliable response evaluation, as overlooking these events can lead to misclassification under RECIST and potentially incorrect assessment of disease progression.

Manual linking can be inconsistent, and lesion diameter calculations are sensitive to topological changes~\cite{shafiei2021ct}. Registration-based pipelines are sensitive to local warp errors and do not explicitly model lesion disappearances or new lesions~\cite{huff2023performance}. Graph formulations that label lesion-level events (persistent, new, disappearing, merging) typically rely on geometric proximity and hard thresholds~\cite{rochman2024graph}. Learning-based matchers (e.g., transformers and self-supervised embeddings) improve appearance modeling but require annotated cohorts and seldom handle merges/splits or unequal counts explicitly~\cite{tang2022transformer,vizitiu2023multi,szeskin2023liver,hering2021whole}.
Recent advances in natural images have reframed correspondence problems using optimal transport (OT) theory. OT provides a principled assignment framework; however, \emph{balanced} OT enforces equal mass and is misaligned with longitudinal scenarios where lesions can emerge, vanish, grow or shrink~\cite{liu2020semantic, de2023unbalanced}. 
\begin{figure}
    \centering
    \includegraphics[width=1\linewidth]{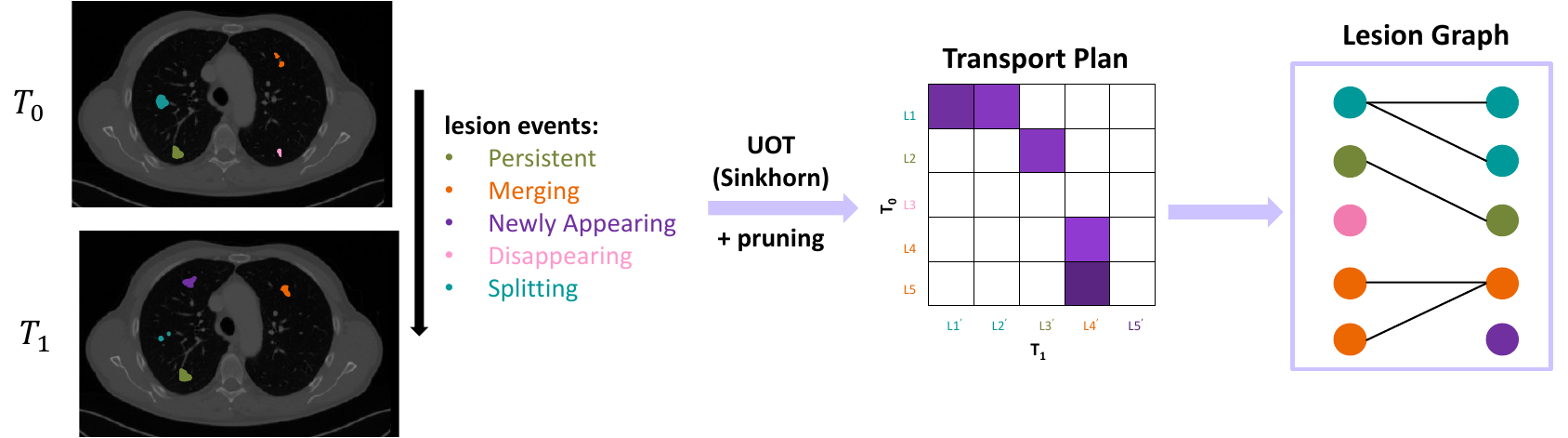}
    \caption{Overview of our Unbalanced Optimal Transport framework for lesion matching.}
    \label{fig:overview}
\end{figure}
We formulate lesion correspondence as \emph{unbalanced} OT (UOT) \cite{de2023unbalanced} over lesion instances, allowing for mass creation and deletion and thereby naturally handling lesion appearance, disappearance, merging, and splitting. Our contributions are threefold. 
(i) To our knowledge, this is the first approach to cast longitudinal lesion correspondence as a UOT problem, providing a principled alternative to distance-based bipartite matchers. 
(ii) We introduce two lesion specific cost modifiers: a registration trust term derived from the deformation field Jacobian, and a local appearance consistency term computed via zero mean normalized cross correlation (ZNCC), which compares the structure between two image patches after removing local intensity bias. These terms adapt the transport cost to spatially varying reliability cues in the images. 
(iii) We propose a patient level tumor load informed asymmetry prior that adjusts the marginal relaxation parameters according to global burden changes, making new or disappearing lesions more or less plausible depending on the clinical context. 
Together, these components produce an interpretable lesion evolution graph with persistent, new, disappearing, merging, and splitting events without requiring heuristics or training data.
We evaluate our lesion matching strategy on synthetic and clinical data, highlighting the benefits of UOT and the added value of our proposed cost terms for longitudinal lesion matching.

\section{Methods}
\label{sec:methods}

\subsection{Problem setup}

Figure \ref{fig:overview} illustrates our proposed method. We consider two CT timepoints for the same patient: baseline $T_0$ and follow-up $T_1$. After non-rigidly registering $T_1$ into the spatial frame of $T_0$, we extract all lesion instances from both timepoints as 3D connected components of the segmentation masks. Let $\mathcal{L}_0$ be lesions at $T_0$ and $\mathcal{L}_1$ be lesions at $T_1$ (already warped). Our goal is to (i) link lesions that represent the same physical finding and (ii) label lesion-level evolution events: \emph{persistent}, \emph{disappearing}, \emph{newly appearing}, and \emph{merging / splitting}.
Each lesion $\ell$ has a binary mask $M_\ell$. We represent it by (a) its centroid $x_\ell \in \mathbb{R}^3$ in physical coordinates and (b) an its sphere equivalent radius $
r_\ell \;=\; \Big(\tfrac{3}{4\pi} \,\mathrm{vol}(M_\ell)\Big)^{1/3}.$

\subsection{Unbalanced Optimal Transport}

We formulate lesion matching as an UOT problem and solve for soft correspondence between lesions at baseline ($T_0$) and follow-up ($T_1$). Let $a_i \ge 0$ and $b_j \ge 0$ denote lesion ``masses,'' proportional to lesion volume and normalized such that $\sum_i a_i = \sum_j b_j = 1$. Unlike classical (balanced) optimal transport, UOT allows mass to be created or destroyed, which naturally models lesions that appear, disappear, grow, or shrink.

We estimate a non-negative transport plan $\Gamma \in \mathbb{R}_+^{|\mathcal{L}_0|\times|\mathcal{L}_1|}$ that specifies how much mass from lesion $i$ is assigned to lesion $j$ by minimizing
\[
\langle \Gamma , C \rangle
+ \lambda \, \mathrm{KL}(\Gamma \mathbf{1} \,\|\, a)
+ \mu \, \mathrm{KL}(\Gamma^\top \mathbf{1} \,\|\, b)
- \varepsilon \, H(\Gamma),
\]
where $\langle \Gamma , C \rangle = \sum_{ij} \Gamma_{ij} C_{ij}$ is the total transport cost, $H(\Gamma)$ is the Shannon entropy, and $\varepsilon>0$ is an entropic regularization parameter. The Kullback–Leibler terms relax the marginal constraints, enabling partial mass transport to account for disappearing or newly appearing lesions. 

\subsection{Cost function for lesion matching}

UOT requires a pairwise cost matrix $C$ that measures dissimilarity between each lesion $i \in \mathcal{L}_0$ and follow-up lesion $j \in \mathcal{L}_1$. We instantiate $C$ with a size-normalized geometric distance and two additional lesion-specific reliability cues.

\textbf{Normalized distance \cite{xia2024deep}.}
The base geometric cost is
\[
c^{\text{geom}}_{ij} =
\frac{\lVert x_i - x_j \rVert_2}{r_i + r_j},
\]
where $x_i$ and $x_j$ are lesion centroids in physical space and $r_i$, $r_j$ are equivalent spherical radii derived from their volumes. This normalization yields a scale-aware measure that avoids bias against small lesions, and the distance is capped to prevent extreme values.

\textbf{Registration trust.}
We estimate local registration reliability using the Jacobian determinant $J(x)$ of the deformation field $\mathbf{u}(x)$ that maps $T_1$ into $T_0$. The local distortion magnitude $e(x) = |\log \tilde{J}(x)|$, with $\tilde{J}(x)$ clipped for numerical stability, is converted into a confidence score $t(x) = \exp(-\beta e(x))$. Averaging $t(x)$ over a dilated lesion mask yields a lesion-level trust $T_\ell$, and we define $\tau_{ij} = \tfrac{1}{2}(T_i + T_j)$ for each candidate pair.

\textbf{Appearance consistency.}
To assess temporal stability, we compute zero mean normalized cross correlation (ZNCC) between patches centered on the lesion centroids in $T_0$ and the registered $T_1$. ZNCC values in $[-1,1]$ are rescaled to $[0,1]$ to obtain per-lesion stability scores $s_\ell$, and for each pair we use $\bar{s}_{ij} = \tfrac{1}{2}(s_i + s_j)$ as the joint appearance reliability.

\textbf{Final cost.}
The complete pairwise cost combines geometric distance with the two reliability cues:
\[
C_{ij} =
c^{\text{geom}}_{ij}
\bigl(1 + w_J [1 - \tau_{ij}] \bigr)
\bigl(1 - w_S \bar{s}_{ij} \bigr),
\]
where $w_J$ and $w_S$ control the influence of registration trust and appearance consistency, respectively. Candidate matches in reliable, appearance-consistent regions are therefore assigned lower costs, while unstable regions remain penalized.

\subsection{Asymmetric marginal weighting}
UOT allows lesions to “appear” or “disappear” by relaxing mass conservation, using penalties $\lambda$ (prevents loss in baseline) and $\mu$ (prevents new mass in follow-up). High $\lambda$ resists disappearing lesions; high $\mu$ resists new ones appearing.

We make these penalties patient-specific using the overall tumor-burden
change
\[
\rho \;=\;
\frac{\sum_{j \in \mathcal{L}_1} \mathrm{vol}(M_j)}
     {\sum_{i \in \mathcal{L}_0} \mathrm{vol}(M_i)} \, .
\]

If $\rho>1$  (disease grew), we lower $\mu$ to ease detection of new lesions.
If $\rho<1$ (disease shrank), we lower $\lambda$ to ease detection of disappearing lesions.
This adaptive weighting helps match lesions as they appear or disappear.



\subsection{Graph extraction}
The plan $\Gamma$ is converted into a lesion-evolution graph by pruning. For each baseline lesion $i$, keep only columns $j$ where $\Gamma_{ij}$ is within a set fraction ($\tau_{\mathrm{row}}$) of that row’s max. For each follow-up lesion $j$, keep only rows $i$ where $\Gamma_{ij}$ is within $\tau_{\mathrm{col}}$ of that column’s max. Remaining $(i,j)$ pairs are predicted edges.
This produces a bipartite graph between baseline lesions and follow-up lesions. Graph node degrees directly reveal lesion-level events:
\begin{itemize}[leftmargin=*]
\item Baseline lesion $i$ has out-degree $0$: \emph{disappearing}.
\item Follow-up lesion $j$ has in-degree $0$: \emph{newly appearing}.
\item If a pair $(i,j)$ is 1--1 (both degrees $=1$): \emph{persistent}.
\item Multiple baseline lesions map into the same follow-up lesion ($\deg(j)\ge2$): \emph{merging}.
\item Single baseline lesion maps to multiple follow-up lesions ($\deg(i)\ge2$): \emph{splitting}.
\end{itemize}
Note that \textit{persistent} here refers only to lesion identity across time; the lesion may still exhibit volumetric growth or shrinkage, which is not modeled in the correspondence graph.

\section{Experiments and Results}
\label{sec:results}

\subsection{Datasets}

{\bf Synthetic data.} To ensure robust evaluation of merging and splitting lesions, we created 30 synthetic cases, each with paired baseline and follow-up volumes with controlled merging, splitting, appearance, and disappearance events.
Each synthetic case consisted of 3D volumes containing $\approx$10 initial lesions at baseline ($T_0$). Strategic lesion placement ensured merging and splitting pairs were positioned for realistic proximity-based fusion. At follow-up ($T_1$), predetermined random transformations allowed the lesions to grow or shrink (persistent), merge, disappear or newly appear.

{\bf Clinical data.} We developed and evaluated on a longitudinal lung CT cohort derived from the released AutoPET~IV challenge dataset (v02)~\cite{Kuestner2025LongitudinalCT}, which includes paired baseline ($T_0$) and follow-up ($T_1$) scans of melanoma patients, with manual lesion annotations from two radiologists. From these, we selected cases with lung lesions and restricted all experiments to this lung subset. After preprocessing and exclusion of unusable cases (e.g., missing deformation fields or corrupt segmentations), we obtained 149 patients in total: 100 for development and 49 held out for testing.

\begin{table}[h]
\centering
\begingroup
\footnotesize
\setlength{\tabcolsep}{4pt}
\renewcommand{\arraystretch}{1.2}
\begin{tabular}{|l|c|c|c|c|c|c|}
\hline
\textbf{Method} &
\textbf{$\mathrm{P}_\text{edge}$} &
\textbf{$\mathrm{R}_\text{edge}$} &
\textbf{$\mathrm{F1}_\text{edge}$} &
\textbf{$\mathrm{wP}_\text{les}$} &
\textbf{$\mathrm{wR}_\text{les}$} &
\textbf{$\mathrm{F1}_\text{topo}$} \\
\hline
Dist-Bipartite &
0.943 &
0.714 &
0.813 &
0.939 &
0.840 &
0.786 \\
\hline
NormDist-Bipartite &
\textbf{1.000} &
0.702 &
0.825 &
0.947 &
0.860 &
0.802 \\
\hline
UOT (ours) &
0.934 &
\textbf{0.795}\textsuperscript{*} &
\textbf{0.859} &
\textbf{0.948} &
\textbf{0.886}\textsuperscript{*} &
\textbf{0.842}\textsuperscript{*} \\
\hline
\end{tabular}
\caption{
Test results. UOT significantly\textsuperscript{*}  outperforms both baselines (one-sided Wilcoxon signed-rank test, $p{<}0.05$).
}
\label{tab:main_results}
\endgroup
\end{table}
\subsection{Evaluation}
We compare against two distance-only baselines:
\emph{Distance-based bipartite}: retains pairs below a centroid-distance threshold and assigns events from vertex degrees (merges allowed).
\emph{Normalized-distance bipartite}: same as above but uses normalized geometric cost with a hard gate to reduce bias against small lesions.
We measure performance at three levels: Edge detection, lesion-state classification, and topology consistency of connected components \cite{rochman2024graph}.

\textbf{(1) Edge detection (precision / recall / F1):}
Each predicted baseline-follow-up correspondence is treated as
an edge; we score it against the reference links.

\textbf{(2) Lesion-state classification (weighted precision / recall):}
Each lesion is labeled as persistent, disappearing, newly-appearing, merging, or splitting. We assign these states from our predicted graph: a follow-up lesion with no incoming matches is “newly-appearing,” and a baseline lesion with no outgoing matches is “disappearing.” We then calculate class-weighted precision/recall for these states. The confusion matrices only use the baseline lesion’s label to avoid double counting, so a missed match appears as NEW, which corresponds to one NEW and one DISAPPEARING in pair space.

\textbf{(3) Connected Components (F1):}
We treat each connected component in the bipartite lesion graph as a
local evolution pattern (e.g.\ stable 1--1 tracking, or a many-to-one
merge). We compare predicted vs.\ reference components and report F1 score.
This captures structural correctness beyond isolated edges.

All hyperparameters
were manually tuned on the development split of the clinical dataset, both for the baselines and our method.

\subsection{Results on synthetic data}
To emphasize merge and split detection, we generated synthetic longitudinal volumes with controlled events and compared baselines to our UOT solver (without patch similarity and Jacobian trust, as these volumes lack real image content).

\begin{figure}
    \centering
    \includegraphics[width=1\linewidth]{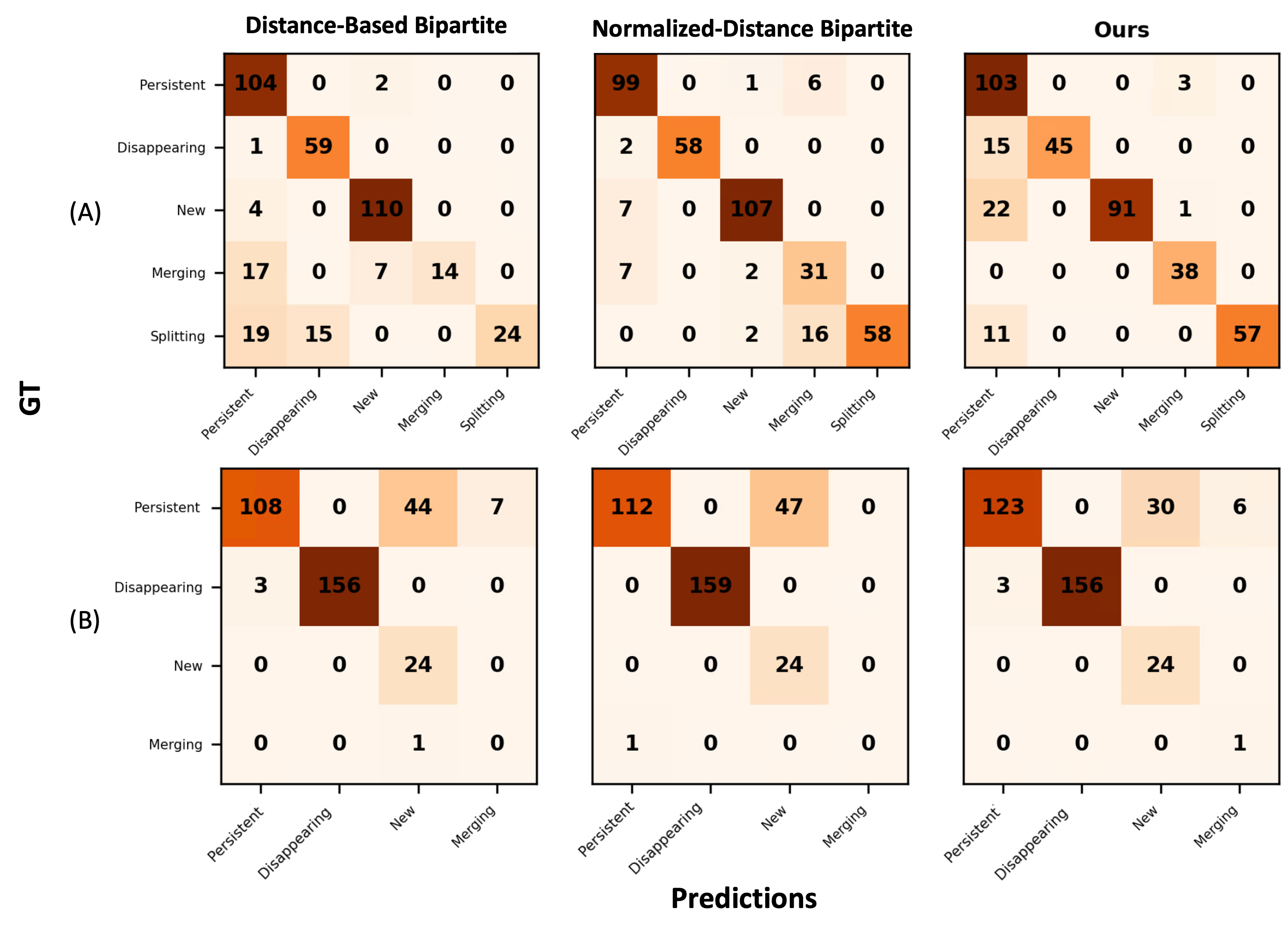}
    \caption{Classification of topology classes on the synthetic dataset (row A) and clinical test set (row B).} 
    \label{fig:conf_toys}
\end{figure}

Fig.~\ref{fig:conf_toys}A highlights the key advantage on the Merging row: UOT correctly places nearly all true merges on the diagonal, whereas distance-only baselines misclassify some as Persistent or New. This reflects UOT’s ability to model many-to-one flow, allowing multiple parents to map coherently to a single child. Detecting merges and splits is critical for accurate burden assessment, as merges can mask growth despite stable or reduced lesion counts, while splits can inflate counts without new disease. 



\subsection{Results on clinical data}
Table~\ref{tab:main_results} shows that UOT surpasses distance-only baselines on most metrics, recovering more true correspondences and producing a lesion-evolution graph closer to the reference. Although precision slightly drops compared to the normalized-distance baseline, the substantial recall gain yields the highest F1-edge overall.

Fig.~\ref{fig:conf_toys}B shows that gains stem from reducing the baselines’ main error: misclassifying persistent lesions as new due to threshold misses. UOT mitigates this by enforcing globally consistent mass flow across larger separations, converting many false ‘new’ calls into correct 1–1 matches. While merges are rare, UOT’s native support for them boosts topology F1 even when multi-parent events are scarce.

Beyond the main comparison in Table~\ref{tab:main_results}, we also carried out a small ablation study to understand how each component influences performance. The UOT alone performs strongly ($\mathrm{F1}_\text{edge}=0.863$), confirming relaxed mass transport as a suitable backbone. Adding the tumor-load-aware asymmetry prior slightly improves lesion-weighted recall (0.892 vs. 0.883), while the patch-based appearance term yields the best edge and topology scores ($\mathrm{F1}_\text{edge}=0.867$), indicating that local texture helps resolve ambiguous geometry. Other cues, such as Jacobian-based registration trust, have minor impact here, likely due to stable underlying registrations. Each component addresses distinct failure modes and is expected to matter more under noisier deformations or larger tumor-load shifts. The full model prioritizes robustness, balancing geometric, appearance, and patient-level cues for consistent correspondences across heterogeneous cases.




\section{Conclusion}
We cast longitudinal lesion correspondence as \emph{unbalanced optimal transport} and enhance it with registration trust, patch-level appearance consistency, and a patient-level tumor-load asymmetry prior. Across held-out patients, this design outperforms distance-only baselines in most evaluation metrics and offers a principled pathway to handle merges and splits within a single, auditable graph formulation.

Future work includes evaluating the method on automatically detected (rather than annotated) lesions and on data with lower registration quality, where larger gains are expected.

\section{Compliance with ethical standards}
This study was conducted retrospectively using open-access human subject data published by K{\"u}stner et al. \cite{Kuestner2025LongitudinalCT}. Ethical approval was not required by the dataset's license.

\section{Acknowledgments}
This project has received funding from the European Union’s Horizon Europe Research and Innovation Programme under grant agreement No. 101095245 ({ONCOVALUE}).

\bibliographystyle{IEEEbib}
\bibliography{strings,refs}
\end{document}